\documentclass[10pt,twocolumn,letterpaper]{article}

\usepackage[pagenumbers]{cvpr} % To force page numbers, e.g. for an arXiv version

\usepackage{graphicx}
\usepackage{amsmath}
\usepackage{amssymb}
\usepackage{booktabs}
\usepackage{soul}
\usepackage{color}
\usepackage{comment}
\usepackage{arydshln}
\usepackage{tikz}
\newcommand{\xmark}{%
\tikz[scale=0.12] {
    \draw[line width=0.7,line cap=round] (0,0) to [bend left=6] (1,1);
    \draw[line width=0.7,line cap=round] (0.2,0.95) to [bend right=3] (0.8,0.05);
}}
\newcommand{\cmark}{%
\tikz[scale=0.23] {
    \draw[line width=1.5,line cap=round] (0.25,0) to [bend left=10] (1,1);
    \draw[line width=1.5,line cap=round] (0,0.35) to [bend right=1] (0.23,0);
}}

\usepackage{xcolor}

\usepackage[pagebackref,breaklinks,colorlinks]{hyperref}

\usepackage[accsupp]{axessibility}

\usepackage[capitalize]{cleveref}
\crefname{section}{Sec.}{Secs.}
\Crefname{section}{Section}{Sections}
\Crefname{table}{Table}{Tables}
\crefname{table}{Tab.}{Tabs.}

\def\cvprPaperID{61} % *** Enter the CVPR Paper ID here
\def\confName{CVPR}
\def\confYear{2023}

\begin{document}

%\title{BinaryViT: What can CNNs teach ViTs about binary neural networks?}

\title{BinaryViT: Pushing Binary Vision Transformers Towards Convolutional Models}

\author{Phuoc-Hoan Charles Le
\thanks{This work was done when Phuoc-Hoan Charles Le was an intern at Huawei Noah's Ark Lab Montreal Research Center.}\\
{\tt\small le.charles55@gmail.com}
\and
Xinlin Li\\
Huawei Noah’s Ark Lab \\
{\tt\small xinlin.li1@huawei.com}
}
\maketitle
%%%%%%%%% ABSTRACT
\begin{abstract}

%{\vahid 1) What is the problem? 2) Why is it important? 3) What the others have done so far in this domain 3) What are the shortages  of the other approaches 4) How did we address these shortages? Each point must be 1 or 2 sentences in the abstract. Follow the same pattern in the Introduction section, each point must be 1 or 2 paragraphs.}

With the increasing popularity and the increasing size of vision transformers (ViTs), there has been an increasing interest in making them more efficient and less computationally costly for deployment on edge devices with limited computing resources. Binarization can be used to help reduce the size of ViT models and their computational cost significantly, using popcount operations when the weights and the activations are in binary. However, ViTs suffer a larger performance drop when directly applying convolutional neural network (CNN) binarization methods or existing binarization methods to binarize ViTs compared to CNNs on datasets with a large number of classes such as ImageNet-1k. With extensive analysis, we find that binary vanilla ViTs such as DeiT miss out on a lot of key architectural properties that CNNs have that allow binary CNNs to have much higher representational capability than binary vanilla ViT. Therefore, we propose BinaryViT, in which inspired by the CNN architecture, we include operations from the CNN architecture into a pure ViT architecture to enrich the representational capability of a binary ViT without introducing convolutions. These include an average pooling layer instead of a token pooling layer, a block that contains multiple average pooling branches, an affine transformation right before the addition of each main residual connection, and a pyramid structure. 
Experimental results on the ImageNet-1k dataset show the effectiveness of these operations that allow a binary pure ViT model to be competitive with previous state-of-the-art (SOTA) binary CNN models.
% To the best of our knowledge, experimental results show that, with a pure ViT architecture, we are the first to be competitive with prior binary CNN models on the ImageNet-1k dataset in terms of accuracy and the number of operations (OPs) by applying these proposed operations.
\end{abstract}

%%%%%%%%% BODY TEXT
\section{Introduction}
\label{sec:intro}

Transformers \cite{attentionisallyouneed} have attracted a lot of attention in natural language processing tasks such as BERT \cite{devlin-etal-2019-bert} and GPT \cite{NEURIPS2020_1457c0d6}. Also, they are gaining a lot of attention in computer vision tasks \cite{dosovitskiy2020vit, pmlr-v139-touvron21a}, since they have been able to outperform most CNNs when being pre-trained on large amounts of data with proper data augmentation and regularization.
%However, its huge model size and its large computational cost create a challenge when deployed to resource-constrained hardware devices.
%Also, compared to CNNs, transformers contain only matrix multiplications which is more friendly to GPUs compared to convolutions since the GPU hardware and its libraries are more optimized for matrix multiplications. Therefore, there are more opportunities to make transformers more efficient than CNNs.

With the rising popularity and the increasing size of vision transformers (ViTs), there has been a rising interest in making them more efficient and less computationally costly to deploy them onto edge devices with limited resources such as smartphones, smart watches, etc.
Therefore, model compression techniques such as quantization \cite{liu2021post, lin2022fqvit, zhang-etal-2020-ternarybert, Qin:iclr22, q8bert, li2022qvit}, distillation \cite{jiao-etal-2020-tinybert, zhang-etal-2020-ternarybert, Qin:iclr22}, pruning \cite{zafrir2021prune, gordon-etal-2020-compressing, sanh2020movement}, etc. have been actively studied to reduce the model size and computational cost of transformers. Among these compression methods, quantization can not only reduce the memory requirement of the model but can also replace expensive floating-point operations with simpler fixed-point operations. An extreme form of quantization is binarization. Binarization of weights and activations can facilitate the use of popcount operations to reduce the model size and reduce computational cost. However, matching the performance of a binary transformer with binary weights and activations with its full-precision counterpart is still a challenging task.

\begin{figure*}
\begin{center}
\begin{subfigure}{0.5\linewidth}
    \hspace*{-4.3cm}
    \scalebox{1.9}{
    \includegraphics[width=\linewidth]{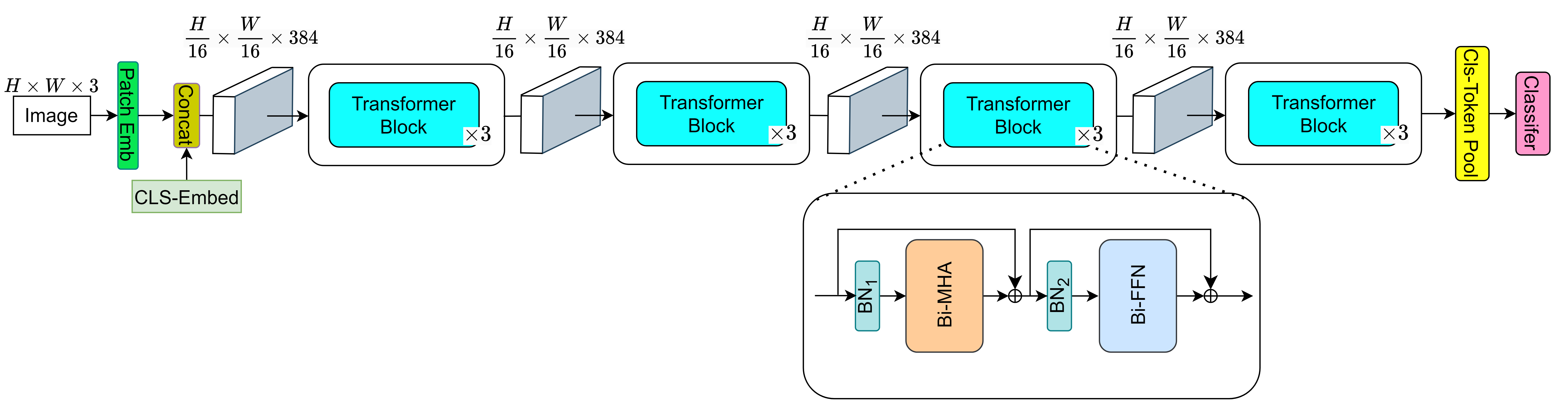}}
    \caption{Baseline binary ViT with the DeiT-S structure}
    \label{fig:baseline-binary-vit}
\end{subfigure}
\end{center}

\hspace*{1cm}

\begin{center}
\begin{subfigure}{.5\linewidth}
    \hspace*{-4.7cm}
    \scalebox{2.0}{
    \includegraphics[width=\linewidth]{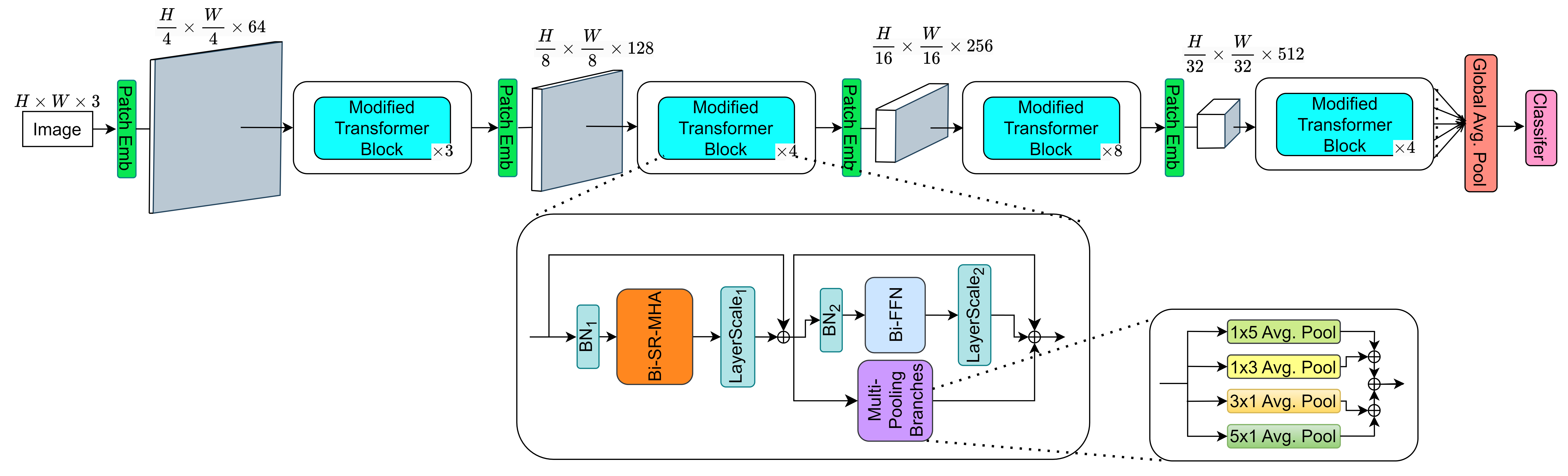}}
    \caption{Our BinaryViT with a global average pooling layer, Multi-Pooling Branches, LayerScale, and a pyramid structure}
    \label{fig:our-binary-vit}
    \label{fig:binary-vit}
\end{subfigure}
\end{center}
\caption{
%{\vahid Visualize better, do not write on box, use visual elements not math in a box.  } {\vahid such a bad figure. having this figure is enough for rejection of the paper. Visualize better or remove it.} 
We start from the baseline binary ViT architecture in (a) and slowly change the architecture to BinaryViT in (b).}
\end{figure*}

Previous works have shown that the drop in the performance can be mitigated with distillation methods from \cite{zhang-etal-2020-ternarybert, Qin:iclr22} to encourage the binary transformers to mimic the full-precision model. Also, a scaling factor can be applied to reduce the drop in the performance of DNNs due to binarization to minimize the quantization error \cite{xnornet}. The scaling factor can also be determined by setting it as a learnable parameter and training it to minimize the task loss \cite{liu2022bit}. For CNNs, \cite{liu2018bi} proposes Bi-RealNet a binary CNN that has extra residual connections to preserve more information to increase its representational capability to make it more accurate on tasks that have lots of classes like ImageNet-1k \cite{imagenet}. \cite{liu2020reactnet} proposes ReActNet which further improves upon the work Bi-RealNet \cite{liu2018bi} by adding a learnable threshold before the sign function and by adding the RPReLU activation function after each residual connection to help reshape the output distribution at near zero extra cost. For MLPs, BiMLP \cite{bimlp} tries to solve the limited representational capability of fully-connected layers by proposing the multi-branch block where they allow patch mixing and channel mixing to happen simultaneously.

However, binary pure ViTs suffer huge performance drop in tasks with large number of classes like ImageNet-1k such that they have lower performance than binary CNNs when applying CNN binarization techniques or applying existing transformer/MLP-based binarization methods onto a pure vanilla ViT architecture since existing binarization methods do not go into details about how architectural or operational designs, besides additional extra residual connections, affect the performance of binary neural networks. 
BiMLP \cite{bimlp} only explores the limited representational capability of a binary fully-connected layer itself and does not explore other architectural details
% that can affect the accuracy of a binary fully-connected-only-based model as this 
as this work does.

Therefore, in this work, after we design a baseline binary pure ViT using existing binarization techniques in Section \ref{sec:baseline}, we analyze the differences between the architecture of a CNN model such as ResNet \cite{he2016resnet} and the architecture of a pure vanilla ViT such as DeiT. From our analysis in Section \ref{sec:method}, we find that binary vanilla ViTs miss out on a lot of key architectural properties that CNNs have that allow CNNs to have much higher representational capability than binary vanilla ViTs. 
Therefore, we propose BinaryViT, in which inspired by the CNN architecture, we include operations from the CNN architecture into a pure ViT architecture in Section \ref{sec:method1}, \ref{sec:method2}, \ref{sec:method3}, and \ref{sec:method4} to enrich the representational capability of a binary pure ViT without introducing convolutions and without significantly increasing the number of operations and the number of parameters. These include an average pooling layer instead of a token pooling layer to account for information from all tokens/patches; a block with multiple average pooling branches to compensate for the loss of the representational capability from a binary fully-connected layer; an affine transformation right before the addition of each main residual branch to prevent the scale of each main residual branch from overwhelming the scale of each main branch; and a pyramid structure to allow binary features to be processed at higher resolution at the early stages without increasing the computational 
complexity.

To the best of our knowledge, these operations or architectural properties introduced in this work have already been explored in previous works relating to full-precision ViTs, but their effects on the accuracy of binary neural networks or on binary ViTs have not yet been explored. Also, to the best of our knowledge, with these operations, we are the first to be competitive with prior binary CNN models on the ImageNet-1k dataset in terms of accuracy and the number of operations (OPs), using a pure ViT architecture. The source code of this work is available on GitHub \footnote{https://github.com/Phuoc-Hoan-Le/BinaryViT}.

\section{Designing a binary ViT baseline}
\label{sec:baseline}

First, we design a baseline binary pure ViT model using the DeiT \cite{pmlr-v139-touvron21a, dosovitskiy2020vit} backbone, since, to the best of our knowledge, there has never been works on a pure ViT model with binary weights and activations. In this section, we use existing binarization techniques applied on CNN, and on BERT \cite{devlin-etal-2019-bert} to design the baseline binary pure ViT model as shown in Figure \ref{fig:baseline-binary-vit} and \ref{fig:bimha}.

\subsection{Binary fully-connected layer}

For the forward pass, for each matrix multiplication, a sign function is used for each input activation matrix, $\textbf{X} \in \mathbb{R}^{N \times D_{in}}$, and weight matrix, $\textbf{W} \in \mathbb{R}^{D_{\mathrm{in}} \times D_{\mathrm{out}}}$. A threshold vector, $\beta_{\textbf{X}} \in \mathbb{R}^{D_{\mathrm{in}}}$, can be applied to the real value inputs right before applying the sign function to allow these inputs to have some distributional shift. For the weights, the threshold, $\mu(\textbf{W}) \in \mathbb{R}^{D_{\mathrm{out}}}$, can be determined by computing the mean value of all elements inside the matrix as in \cite{Qin:cvpr20, Qin:iclr22, Qin2022ijcv}. For the activations, the threshold parameter can be determined using backpropagation to minimize the task loss as in \cite{liu2020reactnet, DBLP:conf/bmvc/XuC19}.
After applying the sign function, a scaling factor, $\alpha_{\mathbf{W}} \in \mathbb{R}$, where $\alpha_{\mathbf{W}} = \frac{1}{n} \|\mathbf{W}\|_{1}$, is applied as in \cite{Qin:iclr22}.
Then, the matrix multiplication output,  $\mathrm{Y}(\textbf{X}) \in \mathbb{R}^{N \times D_{\mathrm{out}}}$, can be calculated using popcount, $\otimes$, as 

\begin{equation}
\mathrm{Y}(\textbf{X}) = \alpha_{\textbf{W}} \ \mathrm{Rsign}(\textbf{X}) \otimes \mathrm{sign}(\textbf{W} - \mu(\textbf{W}))
\label{eq:binary}
\end{equation}
where $\mathrm{Rsign} = \mathrm{sign}(\textbf{X} + \beta_{\textbf{X}})$ as in \cite{liu2020reactnet}.

For each binary fully-connected layer, $\mathrm{BiFC}$, in a binary transformer, we connect a residual connection, $\mathrm{R}$, from the input, $\textbf{X}$, to the output of the linear layer, $\mathrm{BN}(\mathrm{Y}(\textbf{X}))$, as:

\begin{equation}
\mathrm{BiFC}(\textbf{X}) = \mathrm{RPReLU}(\mathrm{BN}(\mathrm{Y}) + \mathrm{R}(\textbf{X}))
\label{eq:birealres1}
\end{equation}
to preserve the information from the previous layer as in \cite{liu2018bi, liu2020reactnet}. $\mathrm{RPReLU}$ activation function proposed by \cite{liu2020reactnet} is used after each residual connection. Also, we replace all layer normalization in the ViT model with batch normalization \cite{batchnorm}, $\mathrm{BN}$, since all linear layers have a normalization layer after it, as in \cite{yao2021leveraging}, to enable faster inference and also faster training compared to layer normalization. The residual function, $\mathrm{R}$, from \cite{liu2020reactnet} can be an identity function if the input and output dimensions are the same. If the input dimension is smaller than the output channel by $n$ times, the residual function will be the input concatenation with itself $n$ times. If the input dimension is larger than the output channel by $n$ times, the residual function will be the average pooling function with a stride of $n$ and a window size of $n$. 
Therefore, $\mathrm{R}(\textbf{X})$ can be defined as

\begin{equation}
\mathrm{R}(\textbf{X}) =
\begin{cases}
    \textbf{X}, & c_{in} = c_{out} \\
    \mathrm{Cat}([\textbf{X},\text{for i in range(n)}], \text{dim=2}), & nc_{in} = c_{out} \\
    \frac{1}{n} \sum_{i=1}^{n} \textbf{X}(:,:, \frac{i-1}{n}d_{in}: \frac{i}{n}d_{in})) & c_{in} = nc_{out} 
\end{cases}
\label{eq:birealres}
\end{equation}
where $\mathrm{Cat}$ is the concatenation function.

Using a straight-through estimator \cite{bengio2013estimating}, we approximate the derivative of sign with respect to an input as 
\begin{equation}
\frac{\partial \ \mathrm{sign}(x)}{\partial x} \approx  
\begin{cases}
    1, & |x| \leq 1 \\
    0, & \mathrm{otherwise}
\end{cases}
\label{eq:ste}
\end{equation}

% However, \cite{liu2018bi, liu2020reactnet} propose an alternative to approximate the derivative of the sign function with respect to the activations as 

% \begin{equation}
% \frac{\partial \ \mathrm{sign}(x)}{\partial x} \approx 
% \begin{cases}
%     2 - 2|x|, & |x| \leq 1 \\
%     0, & \mathrm{otherwise}
% \end{cases}
% \label{eq:stebireal}
% \end{equation}
%{\vahid plot these two functions, and their primitive functions (integrals). Give insights why equation (4) must work better. Try to re-wwrite (4) in the form of absolute value so that (3) and (4) is more comparable.}

\subsection{Binary vision transformer}

\begin{table}
\begin{center}
\scalebox{0.92}{
\begin{tabular}{ ccc }
%\hline
Method & Top-1 (\%)\\
\hline
BinaryBERT \cite{bai-etal-2021-binarybert} & 1.4 \\
\hline
BiBERT \cite{Qin:iclr22}  & 33.5 \\
\hline
BiT \cite{liu2022bit}  & 45.7 \\
+Remove FFN-Distill & 46.5 \\
+SGBERT \cite{ardakani-2022-partially} & 47.4 \\
+ReActNet \cite{liu2020reactnet} & 48.5 \\
%\hline
\end{tabular}}
\caption{
%{\vahid Bad table, Table must speak by iteself, readers are not going to look at - + to figure out what the table mean. You may need to add another column.}
Top 1 accuracy on ImageNet with existing transformer-based binarization methods applied on the binary DeiT-S. 
}
\label{tab:binarybert}
\end{center}
\end{table}

A ViT uses N number of transformer encoder blocks and each transformer block contains one  multi-head attention (MHA) module and one feed-forward network (FFN) module. Initially, in the embedding layer, the image, $\textbf x$ $\in$ $\mathbb{R}^{H \times W \times C}$ gets split up into fixed-size patches, $\textbf{x}_p$ $\in$ $\mathbb{R}^{N \times (P^{2} \cdot C)}$, where $(H,W)$ is the image dimensions, C is the number of channels of the input image, $(P,P)$ is the dimension of each image patch, and $N=HW/P^{2}$ is the number of patches. These patches are then linearly projected by $\textbf{E} \in \mathbb{R}^{(P^{2} \cdot C) \times D}$ before being appended with a  cls-token embedding, $\textbf{x}_{\mathrm{class}} \in \mathbb{R}^D$, and summed with the position embeddings, $\textbf{E}_\mathrm{pos} \in \mathbb{R}^{(N+1) \times D}$, as in Eq. \eqref{eq:embedding} 
%{\vahid for equations use eqref not ref, for this case \eqref{eq:embedding}, remove the number with nonumber command for equations that you are not referring in the text.}

\begin{equation}
\textbf{H}_{1} = [\textbf{x}_{\mathrm{class}}; \ \textbf{x}^{1}_{p} \textbf{E}; \ \textbf{x}^{2}_{p} \textbf{E};...; \ \textbf{x}^{N}_{p} \textbf{E}] + \textbf{E}_\mathrm{pos}
\label{eq:embedding}
\end{equation}
Like in previous works on binarization, the operations in the first layer are always kept in full precision.

%{\vahid Define what do you mean by layer Norm here.} 
%{\vahid too messy index for  $\alpha$, you may need to index with Q, K, V only. The same is true  for $D$, what is the difference between $D$ and $d$? Think about your notation before entering them into your paper.} 
The output of the embedding layer $\textbf{H}_{1} \in \mathbb{R}^{(N+1) \times D}$ then becomes the input for the first transformer block. In each transformer block, the input $\textbf H$ $\in$ $\mathbb{R}^{(N+1) \times D}$ first gets normalized by a pre-batch-normalization layer as $\hat{\textbf{H}} = \mathrm{BN}_{1}(\textbf{H})$.
Then, in the MHA module of the binary transformer with $N_H$ attention heads, the output of that batch-normalization is then used to calculate the query, $\textbf{Q}_{h} \in \mathbb{R}^{(N+1) \times D_h}$; key, $\textbf{K}_{h} \in \mathbb{R}^{(N+1) \times D_h}$; and value, $\textbf{V}_{h} \in \mathbb{R}^{(N+1) \times D_h}$, for each head, $h$, as:
\begin{equation}
\begin{aligned}
\textbf{Q}_{h} = \mathrm{BiFC}_{\mathrm{Q}_h}(\hat{\textbf{H}}_h) \\  
\textbf{K}_{h} = \mathrm{BiFC}_{\mathrm{K}_h}(\hat{\textbf{H}}_h)\\ 
\textbf{V}_{h} = \mathrm{BiFC}_{\mathrm{V}_h}(\hat{\textbf{H}}_h)
\end{aligned}
\label{eq:qkv}
\end{equation}
using Eq. \eqref{eq:binary} and Eq. \eqref{eq:birealres1}, where $D_h = D/N_H$.
We then calculate the attention score as,
$\textbf{A}_{h} = \mathrm{Rsign}(\textbf{Q}_{h}) \mathrm{Rsign}(\textbf{K}_{h}^\top)$.

From \cite{liu2022bit}, to get the binary attention probability matrix of each head, $\textbf{P}_{h} \in \mathbb{R}^{(N+1) \times (N+1)}$ we apply the softmax function on the attention score, $\textbf{A}_{h} \in \mathbb{R}^{(N+1) \times (N+1)}$,
and then apply the round-to-nearest-integer function, $\lceil \cdot \rfloor$,

\begin{equation}
\textbf{P}_{h} = \alpha_{\textbf{P}} \Big \lceil \sigma \Big ( \frac{1}{\alpha_{\textbf{P}}} \ \mathrm{Softmax} \Big( \frac{\textbf{A}_{h}}{\sqrt{D_{h}}} \Big), 0, 1 \Big ) \Big \rfloor
\label{eq:att-prob2}
\end{equation}
where $\alpha_{\textbf{P}} \in \mathbb{R}$ is a learnable scaling factor trained using the method from \cite{Esser2020LEARNED} and the $\sigma(x,r_1,r_2)$ function keeps the output to be between $r_1$ and $r_2$. 
A boolean function from \cite{Qin:iclr22} can also be applied on the attention score, but in Table \ref{tab:binarybert}, we find that it performs 12\% worse than applying the softmax function and then rounding the output.
% , since using the method from \cite{Qin:iclr22} causes a mismatch in the back-propagation between the binary model and the full-precision model, as the gradient of the Bool function is set to 1 when the attention score is between -1 and +1 and 0 otherwise, whereas the gradient of the softmax is not necessarily 1 when the attention score is between -1 and +1 and 0 otherwise.

To get the output of each head, $\mathrm{head}_h \in \mathbb{R}^{(N+1) \times D_h}$, the binary attention probability matrix, $\textbf{P}_{h}$, will be multiplied by the binary value matrix, $\bar{\textbf{V}}_h = \mathrm{Rsign}(\textbf{V}_{h})$. Also, to preserve the information of the query, the key, and the value, we also add the query, the key, and the value to the head output as shown below

\begin{equation}
 \mathrm{head}_h = \mathrm{RPReLU}(\mathrm{BN}_\mathrm{at}(\textbf{P}_h \bar{\textbf{V}}_{h}) + \textbf{Q}_{h} + \textbf{K}_{h} + \textbf{V}_{h})
\label{eq:head}
\end{equation}
and as shown in Figure \ref{fig:bimha}.
%{\vahid what is the difference between $h$ and $\mathbf H$ are they both attention heads? } 
% Note that multiplying the attention probability matrix with the value matrix results in a case where one matrix containing the values, $\{0, 1\}$, is multiplied by the other matrix containing the values, $\{-1, 1\}$ prevents popcount operations from happening. However, replacing the softmax function with a sign function  as in \cite{bai-etal-2021-binarybert} performs much worse as shown in Table \ref{tab:binarybert}. {\vahid try not to complicate the matter, either use $0,1$ either use $\pm 1$ for your binary representation or bring it directly to equation (4) by changing the definition directly.}. Thus, according to \cite{liu2022bit}, the matrix multiplication between $\textbf{P}_h$ and $\textbf{V}_h$ can be efficiently computed with popcount operations as:

% \begin{equation}
% \textbf{P}_h \mathrm{Rsign}(\textbf{V}_{h}) = \frac{(2\textbf{P}_{h}-1) \otimes \mathrm{Rsign}(\textbf{V}_{h}) + \mathrm{Rsign}(\textbf{V}_{h})}{2}
% \label{eq:att-pop-count}
% \end{equation}

The outputs from all heads are then concatenated with each other and used by the fully-connected layer, $\mathrm{BiFC}_\mathrm{O}$, to get the multi-head attention output. A main residual connection is then applied to the output of MHA as
\begin{equation}
\textbf{F} = \mathrm{BiFC}_\mathrm{O}(\mathrm{Cat}(\mathrm{head}_{1},...,\mathrm{head}_{N_{H}})) + \textbf{H}
\label{eq:first-res}
\end{equation}

% Then the residual output, $\textbf{F}$, gets normalized by a 2nd pre-batch-normalization layer, $\mathrm{BN}_2$, and goes through the binary feed-forward network (Bi-FFN) layer which has two fully-connected linear layers, $\mathrm{BiFC}_\mathrm{int}$ and $\mathrm{BiFC}_\mathrm{out}$. The output of FFN can be calculated as 

% \begin{equation}
% \mathrm{BiFFN}(\textbf{F}) = \mathrm{BiFC}_\mathrm{out}(\mathrm{BiFC}_\mathrm{int}(\mathrm{BN}_2(\textbf{F})))
% \label{eq:ffn}
% \end{equation}

% A second residual connection is then applied on the FFN output as
% \begin{equation}
% \textbf{H}_{l+1} = \mathrm{FFN}_{l}(\textbf{F}) + \textbf{F}
% \label{eq:second-res}
% \end{equation}
% in the $l$-th transformer block.
% Then, the output of that residual connection, $\textbf{H}_{l+1}$, gets inputted to the next transformer block.

Then the residual output, $\textbf{F} \in \mathbb{R}^{(N+1) \times D}$, gets normalized by a 2nd pre-batch-normalization layer, $\mathrm{BN}_2$, as $\hat{\textbf{F}} = \mathrm{BN}_{2}(\textbf{F})$ and goes through the binary feed-forward network layer, $\mathrm{BiFFN}$, which has two binary fully-connected linear (BiFC) layers. Finally, a second main residual connection is then applied to the FFN output to get $\textbf{R}$
\begin{equation}
\textbf{R} = \mathrm{BiFFN}(\hat{\textbf{F}}) + \textbf{F}
\label{eq:second-res}
\end{equation}
which gets inputted to the next transformer block.

Like in previous works in binarization, the parameter biases, the parameters at the classifier, operations in the softmax, and normalization layers are kept in full precision.

To improve the performance of the binary ViT, we distill the knowledge of a full precision model to a model with binary weights and activations
by minimizing the soft cross-entropy loss between the student's logit and the teacher's logit as in \cite{liu2020reactnet}.
% by minimizing the soft cross-entropy loss, $\mathrm{SCE}$, between the student's logit $\textbf{y}^{S}$ and the teacher's logit $\textbf{y}^{T}$ as shown in Eq.\eqref{eq:L}
% \begin{equation}
% \mathcal{L}_\mathrm{pred} = \mathrm{SCE}(\textbf{y}^{S}, \textbf{y}^{T}) = -\frac{1}{n} \sum_{c=1}^k \sum_{i=1}^{n} \textbf{y}_{ic}^{T} \mathrm{log} \Bigr(\frac{\textbf{y}_{ic}^{S}}{\textbf{y}_{ic}^{T}} \Bigr)
% \label{eq:L}
% \end{equation}
% where $c$ denotes classes and $n$ denotes the batch size.
We do not use distillation loss for the attention scores and output as the performance will significantly degrade just as shown in \cite{Qin:iclr22, liu2022bit}. Also, we do not distill the FFN outputs from \cite{zhang-etal-2020-ternarybert, bai-etal-2021-binarybert, liu2022bit} as it causes a slight loss of accuracy by 1.2\%. Using the partially-random-initialization method from \cite{ardakani-2022-partially} where we initialize the first patch embedding layer from the full precision model, improves the performance by 0.9\%.

\section{What else do binary CNNs have that binary transformers do not have?}
\label{sec:method}

In Table \ref{tab:binarybert}, using all of the aforementioned techniques in Section \ref{sec:baseline} will only get us 48.5\% top-1 accuracy on ImageNet-1k which is far below the accuracy for most SOTA binary CNNs. Therefore, in this section, we further analyze the details/properties of the CNN architecture that have not been explored in the context of binary neural networks and that can help pure ViTs with binary weights and binary activations improve their representational capability to improve their accuracy on a dataset with a large number of classes without introducing convolutions and without significantly increasing the number of operations and the number of parameters. From Table \ref{tab:compare}, even increasing the number of parameters by either increasing the width or depth of the baseline vanilla binary ViT, it has trouble surpassing SOTA binary CNN models.

\begin{table}
\begin{center}
\scalebox{0.92}{
\begin{tabular}{ c|c }
%\hline
Model & $\mathbb{R}(...)$\\
\hline
DeiT-S & 153,216 \\
ResNet-34   & 71,193,472 \\
\end{tabular}}
\caption{
Element-wise representational capability $\mathbb{R}(...)$ for the binary DeiT-S and for the binary ResNet-34.
}
\label{tab:rep}
\end{center}
\end{table}
\vspace{-0.5cm}
\paragraph{Analysis of representational capability.} First, let's analyze and compare the element-wise representational capability of a binary CNN such as a binary ResNet34, $\mathbb{R}(\text{ResNet-34})$ versus a binary ViT such as a binary DeiT-S, $\mathbb{R}(\text{DeiT-S})$ with both of these models using the ReActNet \cite{liu2020reactnet} design. As in \cite{liu2018bi,bimlp}, we quantify the element-wise representational capability as the number of possible absolute values that each element in a matrix/tensor can have.
%{\xinlin It would be better if you can insert the math definition of representational capability here, just one equation}
We calculate the element-wise representational capability of these models by calculating the element-wise representational capability of the tensor that will be the input for the classifier layer, following the steps from \cite{liu2018bi,bimlp}.

\begin{table*}[h!]
\begin{center}
\scalebox{0.92}{
\begin{tabular}{ cccc|c }
%\hline
Global Average  & Multi-Pooling  & LayerScale & Pyramid & ImageNet \\
 Pooling  & Branches &  & Structure & Top-1 (\%) \\
\hline
% \hline
  \xmark & \xmark & \xmark & \xmark & 48.5\\
% \hline
  \cmark & \xmark & \xmark & \xmark & 56.4\\
% \hline
  \cmark & \cmark & \xmark & \xmark & 60.2\\
%\hline
  \cmark & \cmark & \cmark & \xmark & 61.8\\
%\hline
  \cmark & \cmark & \cmark & \cmark & 67.7 \\
%  \cmark & \cmark & \cmark & \cmark* & 70.6 \\
%\hline
\end{tabular}}
\caption{Results of introducing different operations to the binary ViT model. 
%* denotes that the patch embedding layers in the middle are in full precision.
%{\vahid which one is the baseline? A table must speak by itself, Plot the architecture to see where the embedding layer is, and point out each layer in the architecture so that this table becomes more readble by looing at the architecture figure. }
}
\label{tab:abl}
\end{center}
\end{table*}

From Table \ref{tab:rep}, for a binary DeiT-S, we calculate the element-wise representational capability to be $\mathbb{R}(\text{DeiT-S})=153,216$, whereas for a binary ResNet-34, we calculate the element-wise representational capability to be $\mathbb{R}(\text{ResNet-34})=71,193,472$ which is the order of magnitudes greater than the binary DeiT-S. We believe that the representational capability gap leads to the performance gap between binary ResNet-34 and binary ViT.

\vspace{-0.5cm}
\paragraph{Calculating the representational capability.} A detailed explanation of how the element-wise representational capability was calculated for binary DeiT-S and for binary ResNet-34 can be found in Appendix \ref{app:rep-deit} and \ref{app:rep-resnet}, respectively. 
Generally, from \cite{bimlp, liu2018bi}, one single binary linear/convolutional layer contributes $D_{in} \cdot K^2$ to the element-wise representational capability of an output tensor, where $D_{in}$ is the input dimension, $K$ is the kernel size of the square-shaped weight filter. For an average pooling layer with a kernel size of $2\times2$ and a stride of $2\times2$, the element-wise representational capability of an output tensor would be equal to the element-wise representational capability of an input tensor multiplied by 4 such that $\mathbb{R}(\text{out})=4 \times \mathbb{R}(\text{in})$, since that average pooling layer can be seen as an information aggregation of 4 neighboring patches if we ignore the element-wise division involved in average pooling. For the global average pooling layer, the element-wise representational capability of an output tensor would be equal to the element-wise representational capability of an input tensor multiplied by the number of patches/tokens of that input tensor such that $\mathbb{R}(\text{out})=N \times \mathbb{R}(\text{in})$, where $N$ is the number of tokens/patches since the global average pooling layer can be seen as an information aggregation of all patches/token. For affine transformation such as batch-normalization \cite{batchnorm}, we ignore its effect on the representational capability according to \cite{liu2018bi}.
% {\xinlin It seems your design in sec 3.1, 3.2 and 3.4 are all about improving binary range, is it possible to write a story to link them together ?}
\vspace{-0.5cm}
\paragraph{Increasing the representational capability.} To increase the representational capability of ViT architecture and achieve better binary ViT performance, we proposed three designs. (1) Adding global average pooling before the classifier layer. (2) Adding multiple average pooling branches (3) Bringing Pyramid structure from CNN to ViT. Also, we borrow from the successful design of ResNet and MobileNet and placed an affine transformation before the residual branch to prevent the scale of each main residual branch from overwhelming the scale of each main branch such as the MHA output and the FFN output. With these designs added onto the binary ViT, we name the resulting binary ViT, BinaryViT. While trying to improve the ViT with binary weights and activations, we try to keep the number of parameters and the number of OPs to be around the same as this baseline binary ViT or to be around the same as a ReActNet ResNet-34 \cite{liu2020reactnet}. We define our baseline ViT as the DeiT-S \cite{pmlr-v139-touvron21a} architecture with the applied methods mentioned in Section \ref{sec:baseline}. This baseline has $22.1$M parameters and $1.29\times10^8$ operations (OPs).

\subsection{Global average pooling before classifier layer}
\label{sec:method1}

We noticed that most binary CNNs, such as binary ResNet-34 \cite{he2016resnet} and binary MobileNet \cite{howard2017mobilenets} have an average pooling layer at right before the classifier fully-connected layer, whereas the pure vanilla ViT models such as DeiT \cite{pmlr-v139-touvron21a, dosovitskiy2020vit} have a single cls-token pooling layer before the classifier layer. Using a single cls-token pooling layer prevents the information from all tokens from being taken into account. Knowing from \cite{liu2018bi} that each output token right before the classifier layer has a very limited representational capability
%{\xinlin try to use the definition of "representational capability" rather than "range"}
for binary networks compared to full-precision networks, it would be useful to have information from all tokens being taken into account through global average pooling so that the final classifier layer has more flexibility in adjusting its output during training. After the introduction of the global average pooling into the model right before the classifier, the total element-wise representational capability can be increased by up to $196\times$ since there are 196 patches/tokens in the DeiT-S throughout the whole model.
% , increasing the total element-wise representational capability to $153,216\cdot196=30,030,336$.
% {\xinlin explain what is limited range, you may use the equation from the previous work}

We also remove the cls-token embedding from the model such that in our binary ViT the output from the first embedding layer, $\textbf{H}_{1} \in \mathbb{R}^{N \times D}$ is changed from Eq. \eqref{eq:embedding} to 
\begin{equation}
\textbf{H}_{1} = [\textbf{x}^{1}_{p} \textbf{E}; \ \textbf{x}^{2}_{p} \textbf{E};...; \ \textbf{x}^{N}_{p} \textbf{E}] + \textbf{E}_\mathrm{pos}
\label{eq:embedding2}
\end{equation}
where the position embedding is now $\textbf{E}_\mathrm{pos} \in \mathbb{R}^{N \times D}$.

From Table \ref{tab:abl}, after replacing cls-token pooling with average pooling as shown in Figure \ref{fig:binary-vit}, we received an increase in the performance from 48.5\% to 56.4\% top-1 on ImageNet-1k. Note that from \cite{rw2019timm}, average pooling does not change the performance by a lot for a full-precision ViT. Also, the impact of the average pooling right before the final layer on the number of OPs is negligible.

\subsection{More branches}
\label{sec:method2}

We find that a binary convolution has more representational power than a binary fully connected layer if the number of parameters for the binary convolution and for the binary fully-connected layer are the same. Let's consider a fully-connected layer with a binary weight matrix of size $384\times384$ and a binary convolution weight filter of size $3\times3\times128\times128$. For that binary fully connected layer, the output tensor will have an element-wise representational capability of 384 whereas, for that binary convolutional layer, that output tensor will have an element-wise representational capability of 1152. As we can see, the element-wise representational capability of that binary convolutional layer is $3\times$ larger than the element-wise representational capability of that binary fully connected layer.

This huge element-wise representational capability from the binary convolutional layer allows it to be more flexible in adjusting its output. As a result of this huge output element-wise representational capability per layer, more information will be carried to the final layer through the residual connections, allowing the binary CNNs to adjust their output during training more easily than a binary ViT.

Therefore, inspired by \cite{bimlp}, we decided to add 4 branches right beside each FFN block as shown in Figure \ref{fig:our-binary-vit}. However, the 4 branches contain: an average pooling layer of kernel size $1\times3$, an average pooling layer of kernel size $3\times1$, an average pooling layer of kernel size $1\times5$, and an average pooling layer of kernel size $5\times1$. The average pooling branches that have a kernel size of $1 \times K$ apply the filter along the feature map width, whereas the average pooling branches that have a kernel size of $K \times 1$ apply the filter along the feature map height. The multi average-pooling branches, allow the binary ViT to increase its representational capability without significantly increasing the number of parameters or operations.
%The multi-pooling-branch design can easily be implemented as a matrix multiplication without the duplication of input tensors as required by convolutions, since for average pooling, all channels will have the same filter, which is a property exhibited by MLP-based models such as ResMLP \cite{resmlp} and MLP-Mixer \cite{mlpmixer}. 
% This branch design is not necessarily optimal and how to better design the multi-branch structure to have a better trade-off between the number of operations and accuracy remains an open question.

The multi-branch structure increased the performance of the binary ViT from 56.4\% to 60.2\% in Table \ref{tab:abl}. Also, the number of OPs increased from $1.29\times10^8$ to $1.32\times10^8$, but the number of OPs for the current binary ViT is still below a ReActNet ResNet-34.

\subsection{Scaling right before residual connection}
\label{sec:method3}

We find that networks tested for binarization problems such as ResNet \cite{he2016resnet} and MobileNet \cite{howard2017mobilenets} tend to have a batch-normalization layer \cite{batchnorm} right before being added by a residual connection. We are not sure if that batch-normalization placement helps the signal that is propagating through the residual to be normalized as in \cite{swinv2} or if the affine transformation property of batch normalization
prevents the signal of the main branch to be over-consumed by the residual branch as in \cite{cait,transformer-ln}.

In \cite{transformer-ln}, they showed that with pre-norm architectures like most ViTs, the scale of hidden states grows as we go deeper into the layers of these pre-norm models such that 
\begin{center}
$(1+\frac{l}{2})D \leq \mathbb{E}(||\textbf{H}_l||^2_2) \leq (1+\frac{3l}{2})D$ 
\end{center}
where $l$ is the layer index. 
Therefore, even though we mentioned that an element-wise affine transformation should have no effect on a binary model's representational capability, there is a possibility that the information from any main branches in the deeper layer area could be over-consumed by a residual branch, preventing information from any main branches in the deeper layer area to be passed down in the residual connection, such that the model's representational capability would be lower than what we would expect.
%Therefore, even though we mentioned that an element-wise affine transformation should have no effect on a binary model's representational capability, there is a possibility that the scale of the residual connection could be much larger than the scale of the output of the main branch in the deeper layer area such that the information from any main branches in the deeper layer area could be over-consumed by a residual branch, preventing information from any main branches in the deeper layer area to be passed down properly in the residual connection to the final layer, such that the model's representational capability would be lower than what we would expect.

To test this, we test 3 configurations: one with the residual-post-norm connection (res-post-norm) as in ResNets; one with the residual-post-norm connection plus the pre-norm connection (sandwich-configuration \cite{cogview}); and one with the pre-norm connection plus LayerScale \cite{cait}. 
%The LayerScale in the third configuration will replace the normalization layer that is located right before being added by the residual connection.

From our experiments, using the res-post-norm configuration gets us to 61.4\% top-1, using the sandwich configuration gets us to 61.8\%, and using the third configuration gets us to 61.8\%.  However, having any kind of affine transformation right before the residual connection is better than no affine transformation at all. Using the third configuration, we get increased the performance of the binary ViT from 60.2\% to 61.8\% as shown in Table \ref{tab:abl}. Therefore, for each main residual connection in the transformer in the attention and in the FFN, its main branch would contain an affine transformation such that in our binary ViT Eq. \eqref{eq:first-res} and Eq. \eqref{eq:second-res} are changed to
\begin{equation}
\textbf{F} = \alpha_1 \odot \mathrm{BiFC}_\mathrm{O}(\mathrm{Cat}(\mathrm{head}_{1},..,\mathrm{head}_{N_{H}})) + \beta_1 + \textbf{H}
\label{eq:layerscale1}
\end{equation}
\begin{equation}
\textbf{R} = \alpha_2 \odot \mathrm{BiFFN}(\hat{\textbf{F}}) + \beta_2 + \textbf{F}
\label{eq:layerscale2}
\end{equation}
where $\alpha_1, \alpha_2 \in \mathbb{R}^{D}$ are scaling factors and $\beta_1, \beta_2 \in \mathbb{R}^{D}$ are the bias terms.

From \cite{rw2019timm, cait}, the gain in accuracy for using res-post-norm or LayerScale for a full-precision ViT is very small compared to the gain in accuracy we get for a binary ViT.

\begin{figure}
\begin{center}
\begin{subfigure}{0.5\linewidth}
     \hspace*{0.25cm}
    \scalebox{0.85}{
    \includegraphics[width=\linewidth]{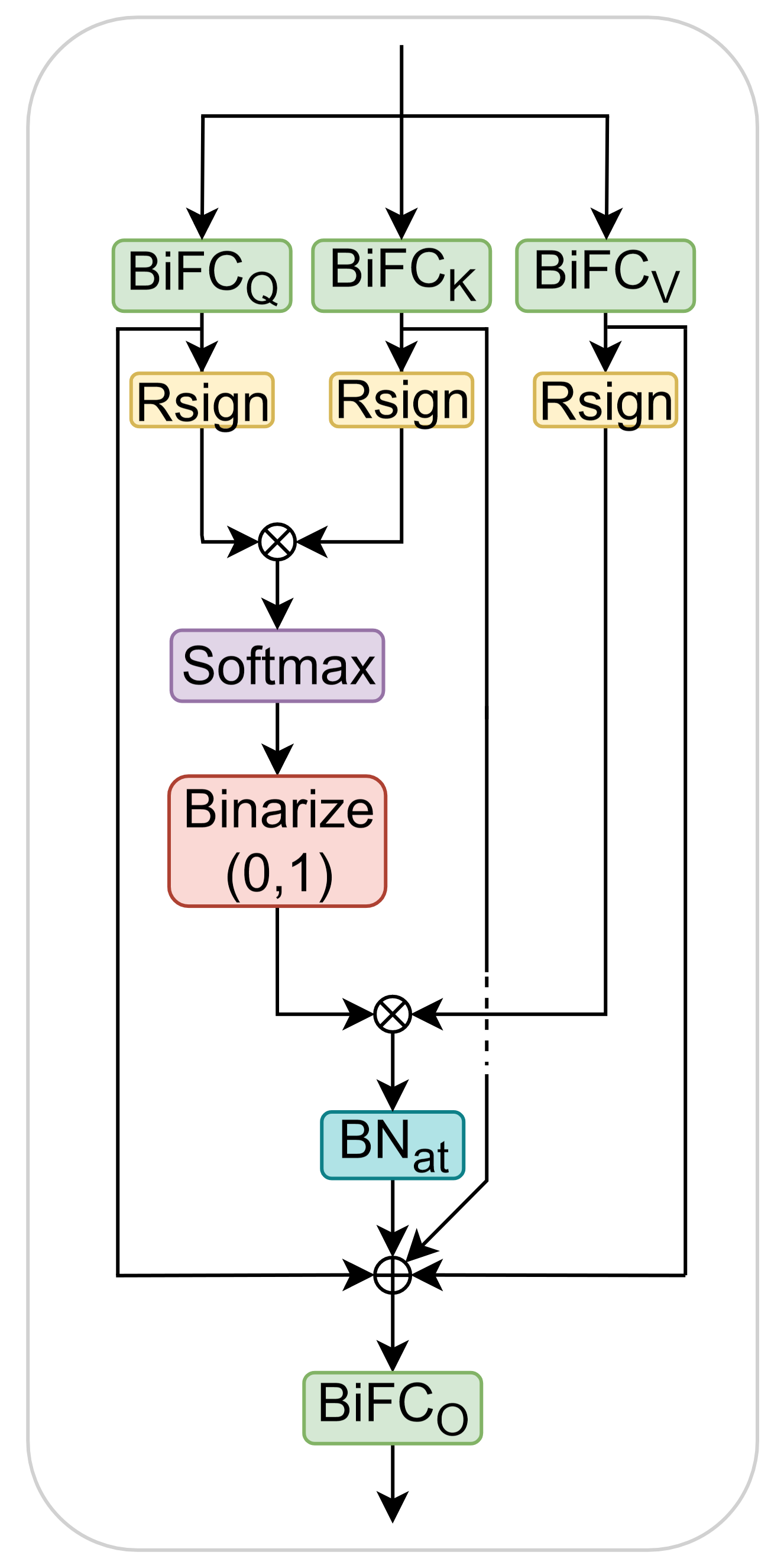}}
    \caption{Bi-Multi-Head-Attention}
    \label{fig:bimha}
\end{subfigure}%
\begin{subfigure}{.5\linewidth}
    \scalebox{1.0}{
    \includegraphics[width=\linewidth]{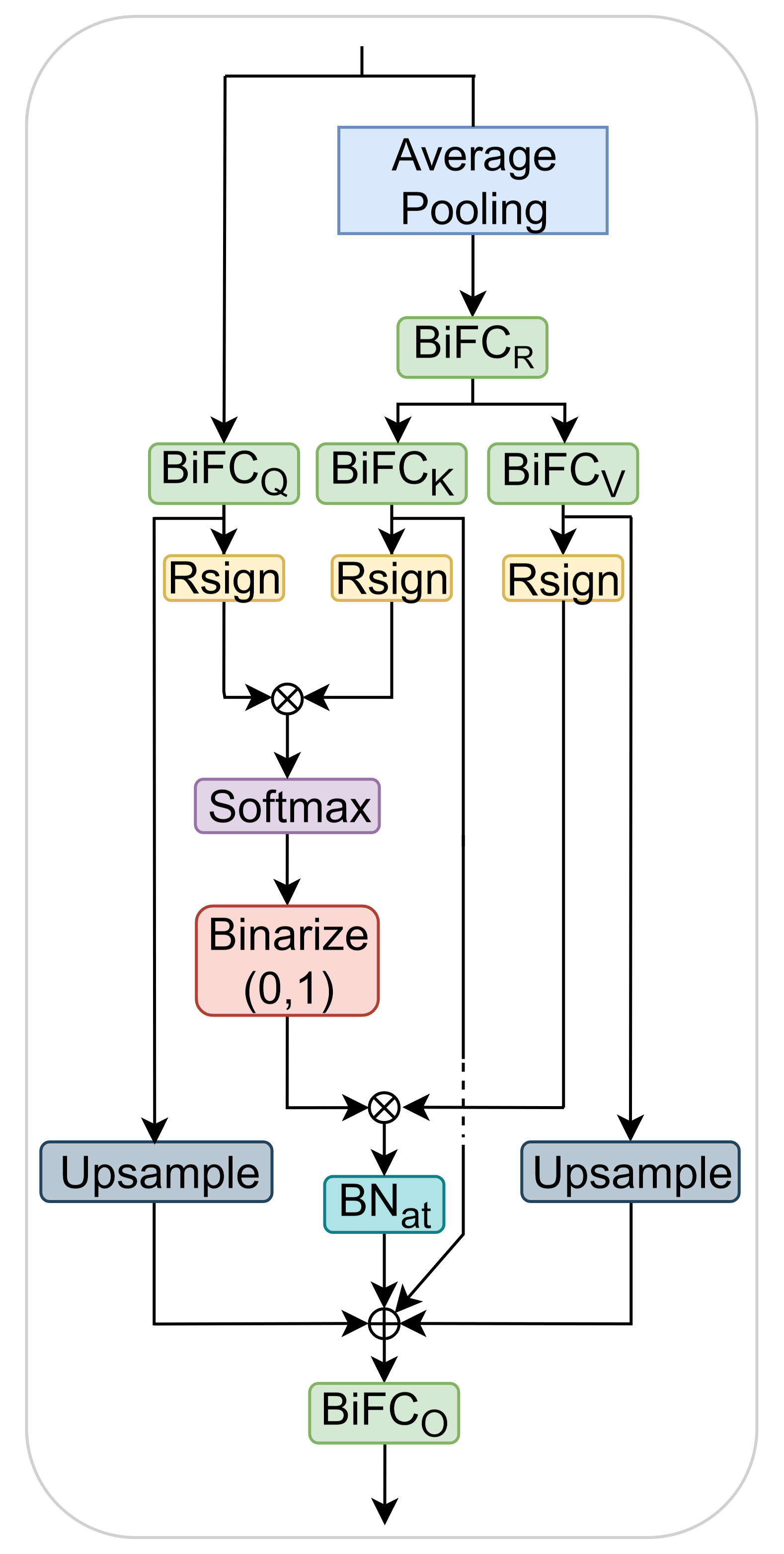}}
    \caption{Bi-Spatial-Reduction Multi-Head-Attention}
    \label{fig:bisrmha}
\end{subfigure}
\label{fig:mha}
\end{center}
\caption{
%{\vahid Visualize better, do not write on box, use visual elements not math in a box.  } {\vahid such a bad figure. having this figure is enough for rejection of the paper. Visualize better or remove it.}  
Normal Bi-Multi-Head-Attention (Bi-MHA) (a) versus Bi-Spatial-Reduction Multi-Head-Attention (Bi-SR-MHA) (b).}
\end{figure}

\subsection{Pyramid structure}
\label{sec:method4}

\begin{table*}[h!]
\begin{center}
\scalebox{0.92}{
\begin{tabular}{c c c c c c c c c c} 
 \hline
 Methods & Model & Global Average  & Multi & Pyramid  & FLOPs & OPs & BOPs & Params & Top-1 Acc \\ 
 & backbone & Pooling & Branches & Stucture & ($\times10^8$) & ($\times10^8$) & ($\times10^9$) & ($\times10^6$) &(\%) \\[0.5ex] 
 \hline\hline
 ReActNet \cite{liu2020reactnet} & ResNet-34 & \cmark & \cmark & \cmark & 1.39 & 1.93 & 3.53 & 21.8 & 67.5\\
 ReActNet-B \cite{liu2020reactnet} & MobileNet & \cmark & \cmark & \cmark & 0.44 & 1.63 & 4.69 & 29.3 & 70.1 \\
 \hline
 BiMLP-S \cite{bimlp} & Wave-MLP-T & \cmark & \cmark & \cmark & 1.21 & 1.56 & \textbf{2.25} & \textbf{17.0} & 70.0 \\
 \hline
 Baseline-S & DeiT-S & \xmark & \xmark & \xmark & 0.57 & 1.29 & 4.51 & 22.1 & 48.5  \\
 Baseline-S$\times$1.5 & DeiT-S$\times$1.5 & \xmark & \xmark & \xmark & 0.57 & 1.70 & 7.15 & 34.9 & 48.7  \\
 Baseline-B & DeiT-B & \xmark & \xmark & \xmark & 1.15 & 3.86 & 17.4 & 86.4 & 60.5  \\
 \hline
 BinaryViT & - & \cmark & \cmark & \cmark & \textbf{0.19} & \textbf{0.79} & 3.83 & 22.6 & 67.7  \\  
BinaryViT* & - & \cmark & \cmark & \cmark & 0.95 & 1.54 & 3.75 & 22.6 & \textbf{70.6}  \\  
 \hline
 \hline
\end{tabular}}
\caption{
%{\vahid Do nto give a wall of numbers. Just select the  important numbers and use bold to emphasize. An ideal table is 5X5. What is the point of having a column 1/1? What does W/A mean? Numbers are not comparable. Make the scale the same, for instance all $10^8$.} 
Comparing results of BinaryViT model with other SOTA binary models with different architecture backbones. * denotes that the patch embedding layers in the middle are in full precision. Floating-point operations (FLOPs) is the number of operations on full-precision layers, bit-operations (BOPs) is the number of operations on binary layers, and from \cite{liu2020reactnet}, $\text{OPs} = \text{BOPs}/64 + \text{FLOPs}$.}
\label{tab:compare}
\end{center}
\end{table*}

Current SOTA binary models with backbones such as ResNet \cite{he2016resnet}, MobileNet \cite{howard2017mobilenets}, and WaveMLP \cite{wavemlp} have a pyramid structure where the feature map size progressively decreases from a high resolution to low resolution and the hidden dimension size progressively increases. However, previous works in binary neural networks overlook the pyramid structure's effect/importance on accuracy.
%Also, BiMLP with a Wave-MLP \cite{wavemlp} backbone uses a pyramid structure, in which all SOTA binary neural networks with good performance have already, but previous works in binary neural networks do not explore the pyramid structure's effect on its performance. 
%However, one work found in SqueezeNet \cite{iandola2016squeezenet} that down-sampling late in the network can improve the accuracy. 

However, we found that the pyramid structure can improve the element-wise representational capability of binary neural networks. Let's consider an example of a network containing 4 stages where each stage contains a different feature map resolution and base hidden dimension like the ResNet model \cite{he2016resnet}, PVT \cite{pvt} model, or the Swin \cite{swin} transformer. One binary fully connected layer with a hidden dimension of 64 in the first stage can contribute up to $64\cdot4\cdot4\cdot4\cdot49=200704$ of the element-wise representational capability, considering the downsampling layers for each transition between stages and the global average pooling layer right before the classifier layer. We can see that this binary fully-connected layer with a low hidden dimension that is applied on a high-resolution feature map can contribute much more to the element-wise representational capability of a binary model than each binary fully-connected layer with a high hidden dimension and a low feature-map resolution such as a binary DeiT-S that has a global average pooling layer, hidden dimension of 384, and a feature map resolution of $14\times14$ in which each of its binary fully-connected layer contributes 75,264 to the model's representational capability. Therefore, we find the pyramid structure can be useful to increase the element-wise representational capability of the model without increasing the computational complexity. 

We design the pyramid architecture to be similar to PVT-S and/or ResNet-34. Our pyramid architecture contains 4 stages with base channel dimensions of 64-128-256-512 as in Figure \ref{fig:our-binary-vit}. The first stage contains 3 identical transformer blocks, the second stage contains 4 transformer blocks, the third stage contains 8 transformer blocks, and the last stage contains 4 transformer blocks. More details on the pyramid architecture can be found in Table \ref{tab:binaryvitstructure}. We designed the pyramid architecture in this way such that we still match the number of parameters of the DeiT-S model \cite{pmlr-v139-touvron21a} and for the ReAcNet method \cite{liu2020reactnet} to be compatible with the transition from the 2nd to the 3rd stage. If the third stage has a base channel dimension of 320, we wouldn't be able to apply the concatenate operation properly as in \cite{liu2020reactnet}, since 320 is not divisible by 128.

For the first and second stages, applying attention to such a large sequence size of 3136 and 784, respectively, is very computationally costly. To remedy this, we apply a down-sampling layer on the input right before the calculation of the key, and value matrix in the self-attention as in the PVT-v2 model \cite{pvtv2} and as shown in Figure \ref{fig:bisrmha}. Therefore, Eq. \eqref{eq:qkv} for the key and value would be changed to 
\begin{equation}
\begin{aligned} 
\textbf{K}_{h} = \mathrm{BiFC}_{\mathrm{K}_h}(\mathrm{BiFC}_\mathrm{R}(\mathrm{AvgPool}(\hat{\textbf{H}}))_h) \\ 
\textbf{V}_{h} = \mathrm{BiFC}_{\mathrm{V}_h}(\mathrm{BiFC}_\mathrm{R}(\mathrm{AvgPool}(\hat{\textbf{H}}))_h)
\end{aligned}
\label{eq:qkv2}
\end{equation}
where $\mathrm{AvgPool}$ is an average pooling function with a kernel size of $R$ and a stride of $R$ and $\mathrm{BiFC}_\mathrm{R}$ is an additional binary linear projection after the average pooling function. Also the calculation of each head from Eq. \eqref{eq:head} would be changed to 
\begin{equation}
\begin{aligned}
 \mathrm{head}_h = \mathrm{RPReLU}(\mathrm{BN}_\mathrm{at}(\textbf{P}_h \bar{\textbf{V}}_{h}) + \textbf{Q}_{h}
             + \mathrm{N}(\textbf{K}_{h}) + \mathrm{N}(\textbf{V}_{h}))
\end{aligned}
\label{eq:head2}
\end{equation}
where $\mathrm{N}$ is the nearest-neighbor interpolation function used to resize the resolution of the key and value feature back to its original resolution after it was downsampled in Eq. \eqref{eq:qkv2}.

In \cite{pvt}, the accuracy gain on ImageNet \cite{imagenet} from introducing a pyramid structure into a full-precision ViT model is negligible, but in Table \ref{tab:abl}, the pyramid structure increased the performance of the binary ViT from 61.8\% to 67.7\% and the number of OPs decreased from $1.32\times10^8$ to $0.79\times10^8$.

\section{Experiments}
% We evaluate our BinaryViT on the ImageNet \cite{imagenet} dataset which contains around 1.2M training images and 50,000 validation images from 1,000 classes. We evaluate our method on ImageNet because most of the SOTA binary CNNs works have been evaluated on that dataset. Our training procedure is described in Appendix \ref{app:exp-settings}.
We evaluate our BinaryViT on the ImageNet \cite{imagenet} dataset because most of the SOTA binary CNNs works have been evaluated on that dataset. The dataset contains around 1.2M training images and 50,000 validation images from 1,000 classes. 
Our training procedure is described in Appendix \ref{app:exp-settings}.
% Appendix \ref{app:exp-settings} describes our training procedure.

\subsection{Experimental Results}
In Table \ref{tab:compare}, we show that our BinaryViT can outperform the baseline binary ViT and 
% even the binary DeiT-B while having significantly less number of OPs and significantly less number of parameters.
even the base version of the baseline binary ViT while having a significantly less number of OPs and a significantly less number of parameters.
This shows the importance of our proposed operations on binary pure ViT models.

We also compare BinaryViT to other SOTA binary models with different architectural backbones such as ReActNet \cite{liu2020reactnet} and BiMLP \cite{bimlp}. 
% From Table \ref{tab:compare}, if we let the down-sampling layers in the patch embedding be run in full precision, our BinaryViT can achieve a competitive performance of 70.6\% top-1 accuracy while having less number of OPs and FLOPs compared to ReActNet CNNs. 
From Table \ref{tab:compare}, BinaryViT can achieve comparable performance with the MobileNet version of the ReActNet-B with having less parameters and less number of OPs, if we let the down-sampling layers in the patch embedding be run in full precision. 
The MobileNet version of ReActNet does not use depth-wise convolutions like the original MobileNet, so it'll have 29.3M parameters. 
Also, BinaryViT can outperform the ResNet-34 version of the ReAcNet, having less number of FLOPs and OPs. 
If the patch embeddings in the downsampling layers are binary, it can outperform the ReAcNet ResNet-34, with having around $7\times$ less number of FLOPs and having around $2\times$ less number of OPs.

Comparing our BinaryViT with BiMLP-S \cite{bimlp}, 
which uses a WaveMLP-T \cite{wavemlp} architecture as its backbone,
we can achieve competitive performance while having a lower number of FLOPs and OPs because our BinaryViT does not use an overlapping convolutional layer in each patch embedding layer which significantly increases the number of FLOPs 
% like in WaveMLP \cite{wavemlp} or 
of the BiMLP. 
Also, the BiMLP is able to achieve competitive performance by just increasing the number of branches, because its backbone architecture, WaveMLP-T, already contains some operations from the CNN architecture, 
such as a global average pooling and a pyramid structure 
in which \cite{bimlp} is unaware of and hasn't explored their importance yet.

Overall, from Table \ref{tab:compare}, we can see that it is very important to have a global average pooling layer, some sort of multi branches layer, and a pyramid structure in which all of the SOTA binary models have in their architecture backbone to have a competitive accuracy.

%When comparing BinaryViT with existing transformer-based binarization techniques, BinaryViT with binary patch embedding layers can outperform these existing transformer-based binarization techniques with $3\times$ less number of FLOPs and $1.6\times$ less number of OPs. The pyramid structure significantly reduces the computational cost of our BinaryViT just like the pyramid structure significantly reduces the computational cost of PVT model \cite{pvt} compared to DeiT model \cite{pmlr-v139-touvron21a}.

\subsection{Benefit of modifying the architecture}

\begin{table}[h]
\begin{center}
\scalebox{0.92}{
\begin{tabular}{c| c c} 
 \hline
 Model & DeiT-S & BinaryViT \\ 
 \hline
 FP32 & 79.9 & 79.9 \\
 1-bit & 48.5 & 70.6 \\
 \hline
 \hline
\end{tabular}}
\caption{
%{\vahid Do nto give a wall of numbers. Just select the  important numbers and use bold to emphasize. An ideal table is 5X5. What is the point of having a column 1/1? What does W/A mean? Numbers are not comparable. Make the scale the same, for instance all $10^8$.} 
Comparisons of the original and our proposed architectures on FP32 and binary settings on ImageNet-1k top-1 accuracy.}
\label{tab:compare-fp}
\end{center}
\end{table}

Table \ref{tab:compare-fp} reports the accuracy gaps between the FP32 and the binary version of the vanilla ViT such as DeiT-S and of the modified ViT. In FP32 setting,
the modified ViT achieves roughly the same performance as the DeiT-S. However, in binary setting, our proposed BinaryViT architecture outperforms the DeiT-S significantly, which justifies the effectiveness of the proposed BinaryViT architecture.

\section{Conclusion}
In this work, we point out that binary vanilla ViTs with backbone architectures such as DeiT miss out on a lot of key architectural properties that CNNs have that allow binary CNNs to have much higher representational capability than binary vanilla ViTs. Thus, we introduce some operations from the CNN architecture into a pure ViT architecture to increase representational capability without the use of convolutions. These include an average pooling layer instead of a token pooling layer, a novel block that contains multiple average pooling branches, an affine transformation right before the addition of each main residual connection, and a pyramid structure. Experimental results on the ImageNet-1k dataset show the effectiveness of the proposed operations to be competitive with prior binary CNNs. 
%There is much more work to be done in keeping the performance of a ViT with binary weights and activations as there is still a huge gap between the accuracy of a full precision ViT and a ViT with binary weights and activations.

%%%%%%%%% REFERENCES
{\small
\bibliographystyle{ieee_fullname}
\bibliography{PaperForReview}
}

\clearpage
\appendix

\begin{table}
\begin{center}
\scalebox{0.92}{
\begin{tabular}{ c|c|c|c }
%\hline
 & Output size & BinaryViT\\
\hline
 &    &  \\
Stage 1 & $\frac{H}{4} \times \frac{W}{4}$  & $\begin{bmatrix}
C_1 = 64 \\
R_1 = 8 \\
N_1 = 1 \\
E_1 = 8
\end{bmatrix} \times 3$  \\
 &    &  \\
\hline
 &    &  \\
Stage 2 & $\frac{H}{8} \times \frac{W}{8}$  & $\begin{bmatrix}
C_2 = 128 \\
R_2 = 4 \\
N_2 = 2 \\
E_2 = 8
\end{bmatrix} \times 4$  \\
 &    &  \\
\hline
 &    &  \\
Stage 3 & $\frac{H}{16} \times \frac{W}{16}$  & $\begin{bmatrix}
C_3 = 256 \\
R_3 = 1 \\
N_3 = 4 \\
E_3 = 4
\end{bmatrix} \times 8$  \\
 &    &  \\
\hline
 &    &  \\
Stage 4 & $\frac{H}{32} \times \frac{W}{32}$  & $\begin{bmatrix}
C_4 = 512 \\
R_4 = 1 \\
N_4 = 8 \\
E_4 = 4
\end{bmatrix} \times 4$  \\
 &    &  \\
%\hline
\end{tabular}}
\caption{
%{\vahid Bad table, Table must speak by iteself, readers are not going to look at - + to figure out what the table mean. You may need to add another column.}
Architectural settings for BinaryViT where $C_i$ is the base hidden dimension, $R_i$ is the reduction ratio of the Bi-SR-MHA, $N_i$ is the number of heads in the attention, and $E_i$ is the expansion ration of the FFN in Stage $i$.
}
\label{tab:binaryvitstructure}
\end{center}
\end{table}

\section{Experimental Settings}
\label{app:exp-settings}
We train the binary ViT from scratch mainly following the hyperparameters from DeiT \cite{pmlr-v139-touvron21a}, using PyTorch \cite{pytorch} and we train the model with Adam \cite{adamw, adam} optimizer for 300 epochs with a batch-size of 512, a weight-decay of 0.0, warmup-epochs of 0, and with an initial learning rate of 5e-4 for the cosine learning rate decay scheduler. Also, we do not train with any data augmentation such as Rand-Augment \cite{randaugment}, random erasing \cite{randomerasing}, stochastic depth \cite{stochasticdepth}, CutMix \cite{cutmix}, Mixup \cite{mixup}, and repeated augmentation \cite{repeated-aug1, repeated-aug2}. However, we still use Random Resize Cropping and horizontal flipping. For knowledge distillation, we use a full-precision DeiT-S \cite{pmlr-v139-touvron21a} as our teacher model. Our optimizer and scheduler are from the Timm library \cite{rw2019timm} and our training loop/pipeline is based on the DeiT repository \cite{pmlr-v139-touvron21a}.

\section{Calculation of representational capability}
Following \cite{liu2018bi,bimlp}, we quantify the element-wise representational capability as the number of possible absolute values that each element in a tensor can have. We calculate the element-wise representational capability of these models by calculating the element-wise representational capability of the tensor that will be the input for the classifier layer or the final layer, following the steps from \cite{liu2018bi,bimlp} and we assume the first layer has binary weights only for simplicity.
\subsection{Binary DeiT-S}
\label{app:rep-deit}
Following \cite{liu2018bi}, for a binary DeiT-S, for the first layer, each output element can have a value range from [0, 195,840], since $P\cdot P \cdot C \cdot \text{max}(x)=16\cdot16\cdot3\cdot255 = 195,840$ where $P$ is the patch size, $C$ is the number of channels, and $\text{max}(x)=255$ is the max value an input image can hold. If we want to have a zero mean, then the range would be [-97,920, 97,920]. Therefore, the current representational capability at that point is 97,920. From \cite{liu2018bi}, normalization wouldn't affect the representational capability, since normalization is just an element-wise affine transformation. For each binary fully-connected layer in the attention, for the query, key, value, and in the final projection layer, the element-wise representation capability for each of these layers is $D=384$ which is the base hidden dimension for DeiT-S. For the matrix multiplication between the binary attention probability and the binary value matrix, the element-wise representation capability for this is $N=196$ which is the sequence length for DeiT-S throughout the whole model. For the FFN, it would be $4\cdot D =4\cdot384=1,536$ for each binary fully connected layer in the FFN. If we calculate the total element-wise representational capability of the binary baseline ViT with DeiT-S backbone, $\mathbb{R}(\text{DeiT-S})$, it would be  $(384\cdot4+196+4\cdot384\cdot2)\cdot12 + 97,920= 153,216$, considering the total number of fully connected layers per transformer block, and the total number of transformer blocks in DeiT-S.

\subsection{Binary ResNet-34}
\label{app:rep-resnet}
Following \cite{liu2018bi}, for a binary ResNet-34, for the first layer, each output element can have a value range from [0, 37,485], since $K_1\cdot K_1 \cdot C \cdot \text{max}(x)=7\cdot7\cdot3\cdot255 = 37,485$ where $K_1$ is the kernel size of the first convolutional layer. If we want to have a zero mean, then the range would be [-18,742, 18,742]. Therefore, the current representational capability at that point is 18,742. After the first convolutional layer, the feature map goes through max-pooling which has no effect on the element-wise representational capability of the model. ResNet-34 has 4 stages with each stage containing different feature map resolutions and hidden dimensions. For stage 1 in ResNet-34, each convolutional layer would have an element-wise representational capability of $3\cdot3\cdot64=576$, considering the kernel size and the hidden dimension of one weight filter. During the transition from stage 1 to stage 2, there will be an average pooling layer in the residual with a kernel size of $2\times2$ with a stride of $2\times2$, which can be seen as an information aggregation of 4 neighboring patches if we ignore the element-wise division involve in average pooling. Therefore, the element-wise representational capability would be multiplied by 4, so the current total element-wise representational capability of the model up to this point can be calculated as $(18,742+576\cdot3)\cdot4=81,880$, considering the number of convolutional layers in that stage.
For stage 2 in ResNet-34, each convolutional layer would have an element-wise representational capability of $3\cdot3\cdot128=1,152$. After the transition from stage 2 to stage 3, the current total element-wise representational capability of the model up to this point can be calculated as $(81,880+1,152\cdot4)\cdot4=345,952$. For stage 3 in ResNet-34, each convolutional layer would have an element-wise representational capability of $3\cdot3\cdot256=2,304$. Therefore, after the transition from stage 3 to stage 4, the current total element-wise representational capability of the model up to this point can be calculated as $(345,952+2,304\cdot6)\cdot4=1,439,104$. For stage 4 in ResNet-34, each convolutional layer would have an element-wise representational capability of $3\cdot3\cdot512=4608$. 
After the feature map is finished getting processed in stage 4, there will be a global average pooling layer that aggregates all the information from the remaining patches before getting processed at the final classifier layer. For a $224\times224$ input image, the feature map resolution at stage 4 of ResNet-34 will be $7\times7$, so the total element-wise representational capability of the binary ResNet-34, $\mathbb{R}(\text{ResNet-34})$, up until the final classifier layer, will be $(1,439,104+4,608\cdot3)\cdot49=71,193,472$.

\end{document}